\pdfoutput=1

\documentclass[11pt]{article}

\usepackage[]{ACL2023}

\usepackage{times}
\usepackage{latexsym}

\usepackage[T1]{fontenc}

\usepackage[utf8]{inputenc}

\usepackage{microtype}

\usepackage{hyperref}

\usepackage{amsmath}
\usepackage{booktabs}
\usepackage{graphicx}
\usepackage{amssymb}

\usepackage{tipa}
\usepackage{caption}
\usepackage{subcaption}
\usepackage{adjustbox}

\definecolor{MatplotlibBlue}{HTML}{1f77b4}
\definecolor{MatplotlibOrange}{HTML}{fb7f11}
\definecolor{MatplotlibGreen}{HTML}{2ca02c}
\definecolor{MatplotlibRed}{HTML}{d62828}

\usepackage{multirow}
\usepackage{fdsymbol}

\usepackage[noabbrev,capitalize,nameinlink]{cleveref}
\crefname{section}{\S}{\S\S}
\Crefname{section}{\S}{\S\S}    
\crefformat{section}{#2\S#1#3}  
\Crefformat{section}{#2\S#1#3}
\crefrangeformat{section}{\S\S#3#1#4--#5#2#6}
\Crefrangeformat{section}{\S\S#3#1#4--#5#2#6}
\crefformat{section}{#2\S#1#3}  
\crefrangeformat{section}{\S\S#3#1#4--#5#2#6}

\newcommand\trm{\texttt{Trm}}
\newcommand\lstm{\texttt{LSTM}}
\newcommand\ptr{\texttt{PtrGen}}
\newcommand\makarov{\texttt{M\&C}}

\title{An Investigation of Noise in Morphological Inflection}

\author{
Adam Wiemerslage${ }^{\diamondsuit}$ 
\quad
Changbing Yang${ }^{\clubsuit}$ 
\quad
Garrett Nicolai${ }^{\clubsuit}$  \quad \\
\textbf{Miikka Silfverberg${ }^{\clubsuit}$  \quad
Katharina Kann${ }^{\diamondsuit}$} \\
${ }^{\diamondsuit}$University of Colorado Boulder \\
${ }^{\clubsuit}$University of British Columbia \\
{\tt \{adam.wiemerslage, katharina.kann\}@colorado.edu}}
\begin{document}
\maketitle
\begin{abstract}
With a growing focus on morphological inflection systems for 
languages where high-quality data is scarce, training data noise is a serious but so far largely ignored concern. We aim at closing this gap 
by investigating the types of noise encountered within a pipeline for truly unsupervised morphological paradigm completion and its impact on morphological inflection systems: First, we propose an error taxonomy and annotation pipeline for inflection training data. Then, we compare the effect of different types of noise on multiple state-of-the-art inflection models. Finally, we propose a novel character-level masked language modeling (CMLM) pretraining objective and explore its impact on the models' resistance to noise. 
Our experiments show that various architectures are impacted differently by separate types of noise, but encoder-decoders tend to be more robust to noise than models trained with a copy bias. CMLM pretraining helps transformers, but has lower impact on LSTMs.
\end{abstract}

\section{Introduction}
Neural morphological inflection has shown impressive results for a huge variety of languages \cite{cotterell2016sigmorphon,kodner-etal-2022-sigmorphon}.
Performance is impressive even for languages with very little supervised inflection data, and often generalizes to unseen lemmas.  
However, the language settings that arguably stand to benefit the most from these tools, those with extremely sparse normalized texts, are less likely to have clean, gold-standard data. 
Despite this, inflection training data noise is rarely addressed nor evaluated in popular benchmarks.
Noise, like incorrect annotations, or mixed dialects or orthographies, can arise in inflection data from web-scraping issues \cite{gorman-etal-2019-weird,mccarthy-etal-2020-unimorph}, human error or changes in writing standards in field data \cite{moeller-etal-2020-igt2p}, or system errors when bootstrapping silver-standard data in an unsupervised fashion \cite{kann-etal-2020-sigmorphon,erdmann-etal-2020-paradigm,wiemerslage2022morphological}.
Unsupervised systems are also prone to \textit{over-regularization}, where the dataset contains few or no irregular samples. Datasets derived from FSTs or textbook examples can also display over-regularization \cite{vylomova-etal-2020-sigmorphon}.

In this work, we investigate the impact of noise on inflection generation systems. 
We build an automatic pipeline for annotating inflection noise
and explore the noise distribution that arises in an unsupervised system for bootstrapping inflection data.  
We measure the impact of noise on several state-of-the-art neural inflection generation systems that we benchmark on the SIGMORPHON 2017 shared task development set \cite{cotterell2017conll}. Finally, we explore a novel character-level masked language modeling (CMLM) pretraining objective to mitigate the impact of noise during training.
By this, we aim to shed light on which architectures and training methods are how robust to noise, which types of inflection noise should be targeted in filtering approaches, and how conservatively unsupervised systems should sample inflection pairs.

We find that noise related to slot alignment issues is more common, but also less impactful than noise related to paradigm induction issues.
Architectures with an inductive bias towards copying from the lemma are more effective on datasets that lack sample diversity, but more typical encoder-decoder models are more robust to noise. 
Standard encoder-decoders display better performance on noisy data when pretrained with CMLM, which is especially effective for the Transformer. 
Our code and data are publicly available.
\footnote{\url{https://github.com/Adamits/morphological-inflection-noise}}

\section{Noisy Training Data}
Our data comprises four languages: Icelandic, German, Swedish, and Russian. We describe the source of training data below.

\paragraph{tUMPC}
We focus on training data based on \citet{wiemerslage2022morphological}, who propose a system for truly unsupervised morphological paradigm completion (tUMPC). 
Starting with the Bible corpus \cite{mccarthy-etal-2020-johns} for a given language, tUMPC first clusters data into paradigms using the system from \citet{mccurdy-etal-2021-adaptor}. 
Second, paradigms are clustered into parts-of-speech. Finally, tUMPC aligns similarly inflected forms across paradigms that belong to the same POS.
The result is a dataset of paradigms for which each type is marked with its inflectional slot.
We take all possible pairs of words from the tUMPC paradigms to form inflection training data. 
This dataset is likely to contain noise due to errors in the learning process. In \cref{sec:annotation}, we discuss this noise in detail. 

\paragraph{UniMorph}
tUMPC relies on frequency thresholds to find productive inflection transformations, which makes it likely that morpho-phonological variations and irregular inflections may not be well-attested in the data. 
To control for the potential lack of sample diversity in our training data, we create a second training dataset wherein all samples marked as correct are replaced by a pair sampled from UniMorph \cite{sylak2015universal,batsuren2022unimorph}, a database of morphological paradigms covering hundreds of languages. We sample pairs from the same MSDs as the tUMPC correct samples. 
Following recent results \cite{liu-hulden-2022-transformer,goldman-etal-2022-un,kodner-etal-2022-sigmorphon}, we also ensure that the lemma overlap with the evaluation set is the same as in the original tUMPC data. The resulting dataset has higher lemma diversity, without introducing missing MSDs, and maintains lemma overlap with the evaluation set. 
We combine this data with the noisy training data from tUMPC to investigate the impact of noise when sample diversity is less pervasive.

\begin{table}[t]
    \centering
    \small
    \begin{tabular}{l|rrrr}
        \toprule
         & Deu & Isl & Rus & Swe  \\
         \midrule
         Original & 40956 & 32225 & 128326 & 59814 \\
         Annotated & 6444 &	4735 & 14171 & 6054 \\
         \bottomrule
    \end{tabular}
    \caption{Size of training set for each language before and after filtering out pairs that cannot be annotated.}
    \label{tab:baseline_training_data}
\end{table}

\paragraph{Slot Mapping}
\label{sec:slot-mapping}
tUMPC slots are arbitrary identifiers with no grammatical meaning. 
In order to compare tUMPC to UniMorph, and because we evaluate on a benchmark sampled from UniMorph, we need to map each tUMPC slot to a unique UniMorph slot. UniMorph slots are bundles of morphological tags that express categories like person or tense. 
We refer to these bundles as morpho-syntactic descriptions (MSDs). 
Since the tUMPC training pairs have low overlap with UniMorph, we require a morphological analyzer to get ground truth morphological tags of tUMPC types. 
Here, we use morphological analyzers from Apertium \cite{forcada2011apertium}, which we map to UniMorph MSDs. 
Then, we apply the mapped Apertium to all tUMPC types, resulting in ground truth MSDs for every training sample in tUMPC. We also want the slots predicted by tUMPC expressed as UniMorph MSDs during training. 
For this, we compare the unsupervised slot and gold MSD of all tUMPC types and find the alignment that maximizes their overlap following the evaluation metric from \citet{jin2020unsupervised}. 
Every training sample can then be assigned the (noisy) MSD from this alignment. For more details see Section \ref{sec:appendix_slot_map} in the appendix.

Many inflection pairs created by tUMPC could not be reliably annotated by our pipeline and are thus filtered out.
Table \ref{tab:baseline_training_data} shows the amount of training data before and after filtering.
These are inflection pairs that are not in the Apertium lexicon, but are also not determined to be noise. We additionally filter pairs containing a form with an Apertium analysis that we could not reliably map to UniMorph due to disagreements between the two resources.
We compare the distribution of morphological tags for our dataset according to Apertium before and after applying this filtering step, and find no systematic difference.
We conclude that removing forms due to mapping errors does not bias our data towards certain inflections. These filtering steps remove a majority of data, but still leave reasonably sized training sets for each language.

\section{Related Work}
\paragraph{Morphological Inflection}
Morphological inflection is the task of generating a word form given a lemma and target MSD. For example, given the verb \textit{laugh} and the MSD expressing past tense, the goal is to generate \textit{laughed}. 
Encoder-decoder neural approaches have largely dominated morphological inflection in recent years \cite{faruqui2016morphological,kann-schutze-2016-single}, where the the full target string is decoded from a neural network. 
But neural models that bias the task towards transducing input strings have been shown to be successful in low-resource data settings \cite{aharoni-goldberg-2017-morphological,makarov2017align,makarov-clematide-2018-imitation,sharma2018iit}. 
Shared tasks on morphological inflection \cite{cotterell2016sigmorphon} have spurred large interest in the task, and also serve as an evaluation benchmark.

\paragraph{Learning from Noisy Data}
Several approaches have been proposed for mitigating the impact of noise in machine learning, for example: confidence weighting \cite{rebbapragada2007class}, loss correction \cite{patrini2017making}, and noise-contrastive estimation \cite{gutmann2010noise}. 
For a recent survey on noise robust neural networks, see \citet{song2022learning}. 
Most approaches consider classification, framing noise as label corruption. However, there is also work exploring noise in tasks like machine translation \cite{khayrallah-koehn-2018-impact,michel-neubig-2018-mtnt}, as well as morphological disambiguation \cite{zalmout-etal-2018-noise}. 
We focus on morphological inflection, a conditional generation problem.

\paragraph{Morphological Inflection with Training Noise}
Morphological inflection with noisy data has largely been unexplored.
\citet{moeller-etal-2020-igt2p} report gains in performance after manually cleaning training data that were bootstrapped from interlinear glossed texts.
\citet{nicolai-silfverberg-2020-noise} find that exposing an inflection model to its own mistakes leads to better generalizability.
Our work explores the impact of different types of noise that occur in the training data in detail.

\section{Noise Taxonomy}
\label{sec:annotation}
We develop an automated pipeline to annotate each inflection pair from a taxonomy of noise, primarily relying on rule-based morphological analyzers for each language from Apertium. Here we describe each type of noise in our taxonomy, which we organize into three categories: lemma errors, paired errors, and MSD errors. For a description of how each type of noise is detected, see Section \cref{sec:appendix_detecting_noise} in the appendix. 

\subsection{Lemma Errors}
We first describe noise in which a training sample includes lexical items that should not be in the inflection training data at all. Lemma noise arises due to issues inherent to the corpus (e.g. misspellings, etc), or issues in which lexical items that were sampled from the corpus (e.g. punctuation is not inflection data). We describe two types of noise that fall into this category.
\paragraph{Lexicon Noise}

Any word type that is not in the standard vocabulary
of a given language is considered lexicon noise.
We expect this noise to come from archaisms, borrowings, and biblical references. This means that lexicon noise \textit{could} follow the regular inflections of a language, and in that case may have a low impact on downstream inflection systems. It could also have a positive impact by increasing training data size, likely with higher lemma diversity. However, lexicon noise could also entail archaic or borrowed \textit{inflections}, which would introduce non-existent transformations. In general, lexicon noise could also occur due to language, dialect, or orthography mixing -- though we expect this to be somewhat rare in the Bible corpora that we build upon.

\begin{figure}[t]
     \centering
     \includegraphics[width=0.55\linewidth]{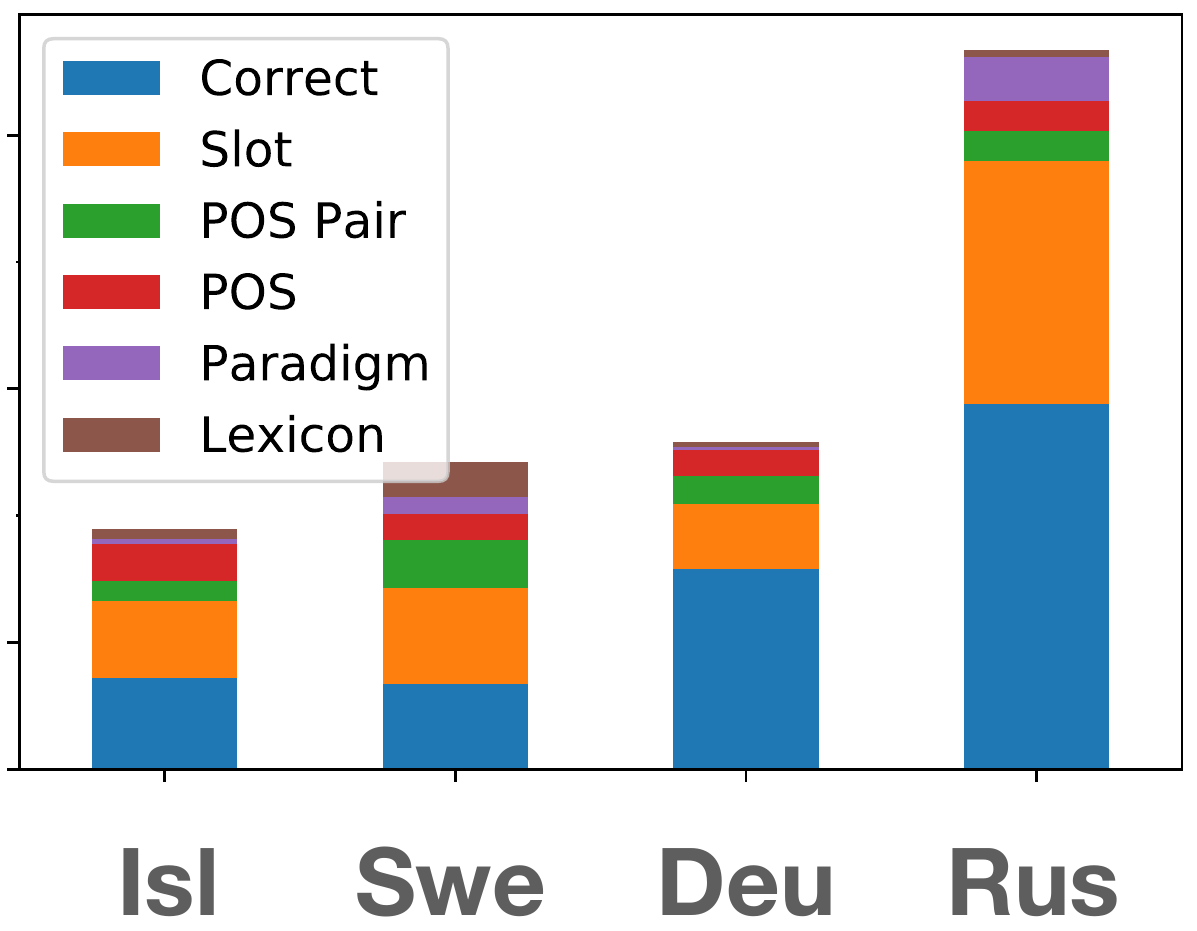}
    \caption{tUMPC noise distribution for all languages.}
    \label{fig:baseline_noise}
\end{figure}

\paragraph{POS Noise}
Not all parts of speech inflect, and this varies by language. However, tUMPC sometimes induces spurious inflection pairs from words that do not inflect, like conjunctions in most languages. Any word of a POS that does not inflect in the given language is thus considered POS noise. 
This is detected with the Apertium POS, but ultimately these samples will be assigned an MSD from UniMorph with a POS that does inflect. See Table \ref{tab:appendix_valid_pos} for the POS that inflect according to this study.

While inflection pairs with POS noise may sometimes have real transformations, it is highly likely that these add spurious inflections to the training data. For this reason, we expect POS noise to consistently cause mistakes in each system.

\subsection{Paired Errors}
Next we describe noise types in which valid lexical items erroneously form an inflection pair. These types of noise occur due to issues in the paradigm induction algorithm that produced the inflection data.

\paragraph{POS Pair Noise}
Two words from different POS forming an inflection pair constitutes POS pair noise. This can occur when tUMPC puts two completely unrelated words in the same paradigm, but this could also arise due to valid derivations. The latter case could be considered a morphological transformation that our model should learn. There is debate around whether or not there is a clearly defined distinction between inflection and derivation \cite{haspelmathinflection}, 
but since our goal is to evaluate on SIGMORPHON shared task data that does not include derivation, we consider this noise in our pipeline. 
POS pair noise should have a similar impact as POS noise, except that due to derivations we expect several of the induced transformations to appear much more commonly in languages where productive derivation is pervasive.

\paragraph{Paradigm Noise}
Pairs of forms that do not belong to the same paradigm but share a POS constitute paradigm noise. An example in English is
\textit{warp} $\rightarrow$ 
\textit{wraps}, where \textit{wraps} 
comes from a different paradigm than \textit{warp}.
Because paradigm noise should contain target words expressing a valid inflection, it is possible that it will not have much negative impact on an inflection system's decoder. However, since paradigm noise also contains source forms that the target is not actually inflected from, the full transformation from source to target has the potential to be spurious. For instance, in the \textit{warp} $\rightarrow$ 
\textit{wraps} example above, consider the apparent \textit{a, r} metathesis. This could cause models relying on a bias towards transduction to struggle more with paradigm noise than models relying predominantly on a decoder.

\subsection{MSD Errors}
Finally we describe noise 
consisting of assigning the wrong MSD to an inflection. We have only a single noise type here. It occurs due to issues in the slot alignment algorithm.

\paragraph{Slot Noise}
When the target word in an inflection pair has an incorrect MSD, we mark this as slot noise. For example, consider an inflection 
\textit{cry} $\rightarrow$
\textit{cried}, which is a valid inflection pair, but if, for example, \textit{cried} is incorrectly marked as the third person present, this would be slot noise. Our training setup makes use of \textit{only} the target slot -- if only the source form has an incorrect slot then a pair is not marked as slot noise.
This error results in a mismatch of MSDs and inflectional transformations. So, for some paradigms, the output form for an inflection can be thought of as swapped with another output form in the same paradigm. If these two forms are different, this is likely to confuse the inflection system by presenting counter-evidence to the correct inflection. Additionally, it is possible that there are few to no correctly tagged instances of rare or irregular inflection classes, causing a system to confidently learn that, e.g., the past tense inflection is the third person present tense.

\subsection{Analysis}

Figure \ref{fig:baseline_noise} presents the noise distribution in the training data for each language according to our annotation pipeline. In German and Russian, there is more correct data than noise, but in Icelandic and Swedish, there is more noise than correct data.
We can also consider noise by its 
source of error in tUMPC: either resulting from an error in slot alignment, 
i.e. slot noise; an error in the corpus, i.e. lexicon noise; or a paradigm induction error, 
i.e. all other noise. Slot alignment issues are the most common source of noise, though, for German and Swedish, there are nearly as many paradigm induction errors. 
More specifically, POS pair noise is the most common noise type after slot noise, and lexicon noise is the least common, most of which occurs in Swedish.

\section{Experiments}

\subsection{Models}
We compare four neural inflection generation systems. 
We implement a bidirectional LSTM with attention \cite[\lstm]{kann-schutze-2016-single}, a Transformer \cite[\trm]{wu-etal-2021-applying}, a pointer generator LSTM \cite[\ptr]{sharma2018iit}, and we use the Dynet implementation of \cite[\makarov]{makarov-clematide-2018-imitation}: a transducer optimized with minimum risk training. 
For all other models, our implementation is based on yoyodyne,\footnote{\url{https://github.com/CUNY-CL/yoyodyne}} which is built on pytorch \cite{NEURIPS2019_9015}. 
All models follow the hyperparameters reported in the original papers, with minor increases in epochs to ensure that they converge. 
Explicit hyperparameters are listed in the appendix in Table \ref{tab:appendix_hyperparams.}.

This gives us two classes of models: general encoder-decoders (\lstm{} and \trm{}), which may struggle for low-resource scenarios and under a lack of sample diversity; and transducer-like models with a bias towards copying from the lemma (\ptr{} and \makarov{}), which are known to perform better in low-resource scenarios by relying on modeling character transduction.
For all results that we report, we train five of each model on the same five random seeds and report the mean.

\subsection{Evaluation}
We evaluate on development sets from the SIGMORPHON 2017 shared task \cite{cotterell2017conll}. 
During training, we reuse the target MSDs found through the mapping described in \Cref{sec:slot-mapping}, which match the SIGMORPHON target MSDs. 
Though our training setup considers \textit{reinflection}, we evaluate on inflection from a lemma. 
This small mismatch in task has a minor negative impact on accuracy \cite{cotterell2016sigmorphon}.

\begin{table}[t] 
\centering 
\small
\begin{tabular}{l|ccccc} 
\toprule 
Lang & \makarov & \lstm & \ptr & \trm \\ 
\midrule
Deu & 28.38 & 23.14 & \textbf{31.34} & 21.42 \\
Isl & \textbf{21.90} & 16.60 & 19.03 & 16.27 \\
Rus & \textbf{38.92} & 31.20 & 33.07 & 32.67 \\
Swe & \textbf{35.88} & 22.34 & 26.05 & 25.68 \\
\midrule
Avg & \textbf{31.27} & 23.32 & 27.37 & 24.01 \\
\bottomrule 
\end{tabular}
\caption{Accuracy for the tUMPC training data.} 
\label{tab:baseline_res.} 
\end{table}

\begin{table}[t] 
\centering
\small
\begin{tabular}{l|ccccc} 
\toprule 
Lang & \makarov & \lstm & \ptr & \trm \\ 
\midrule
Deu & 40.12 & \textbf{45.98} & 42.06 & 34.75 \\
Isl & 31.28 & 28.64 & 30.00 & \textbf{32.00} \\
Rus & 61.80 & 65.58 & 61.96 & \textbf{67.23} \\
Swe & \textbf{54.20} & 49.32 & 47.57 & 53.43 \\
\bottomrule 
Avg & 46.85 & \textbf{47.38} & 45.40 & 46.85 \\
\bottomrule 
\end{tabular}
\caption{Accuracy for the UniMorph training data.} 
\label{tab:resampled_res.} 
\end{table}

\begin{figure*}[t]
     \centering
     \begin{subfigure}[b]{0.46\linewidth}
         \centering
         \small
         \includegraphics[scale=0.26]{
            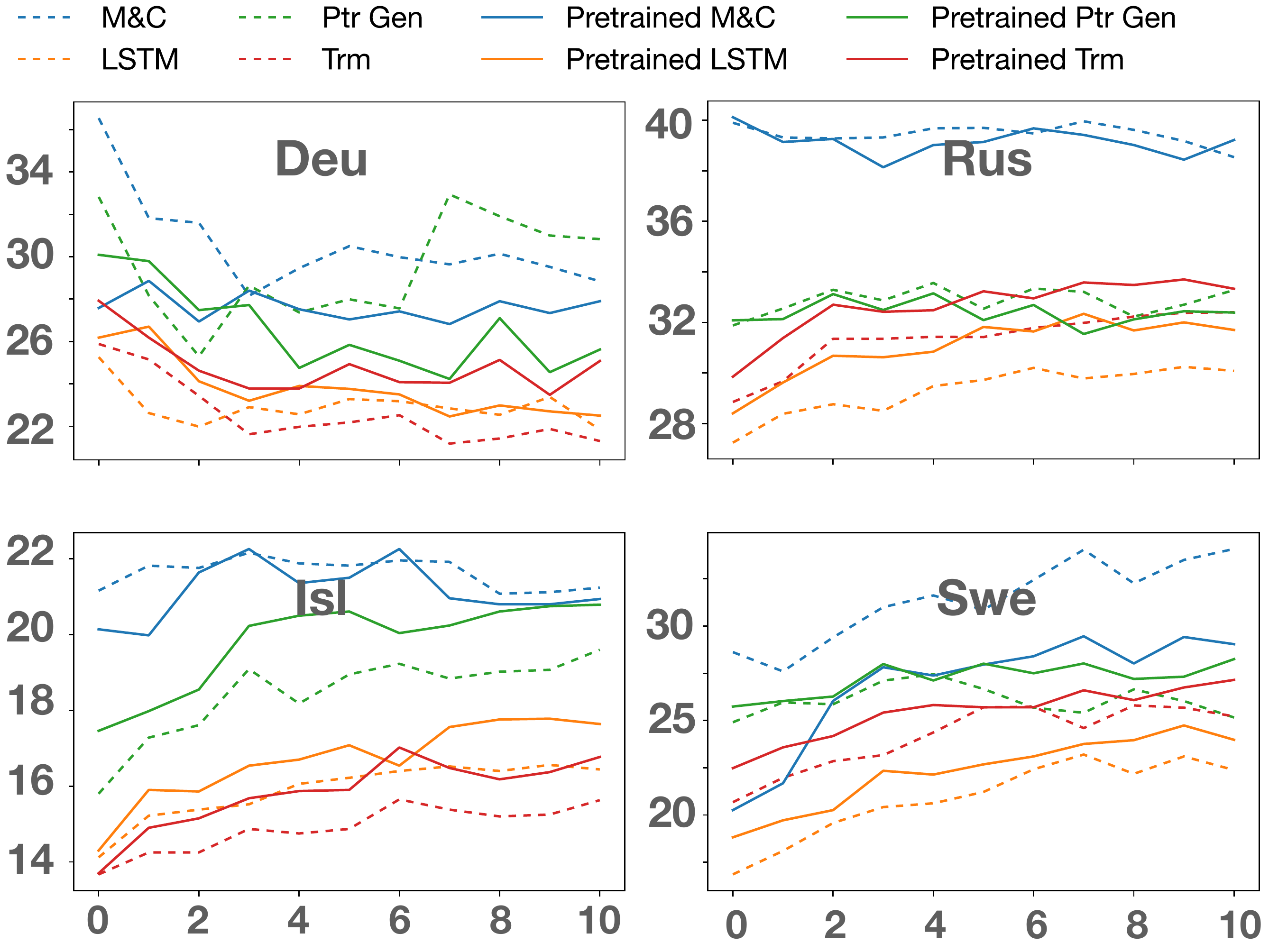}
        \caption{
            tUMPC
        }
        \label{fig:augmented_all_langs_baseline}
    \end{subfigure}
    \begin{subfigure}[b]{0.46\linewidth}
         \centering
         \small
         \includegraphics[scale=0.26]{
            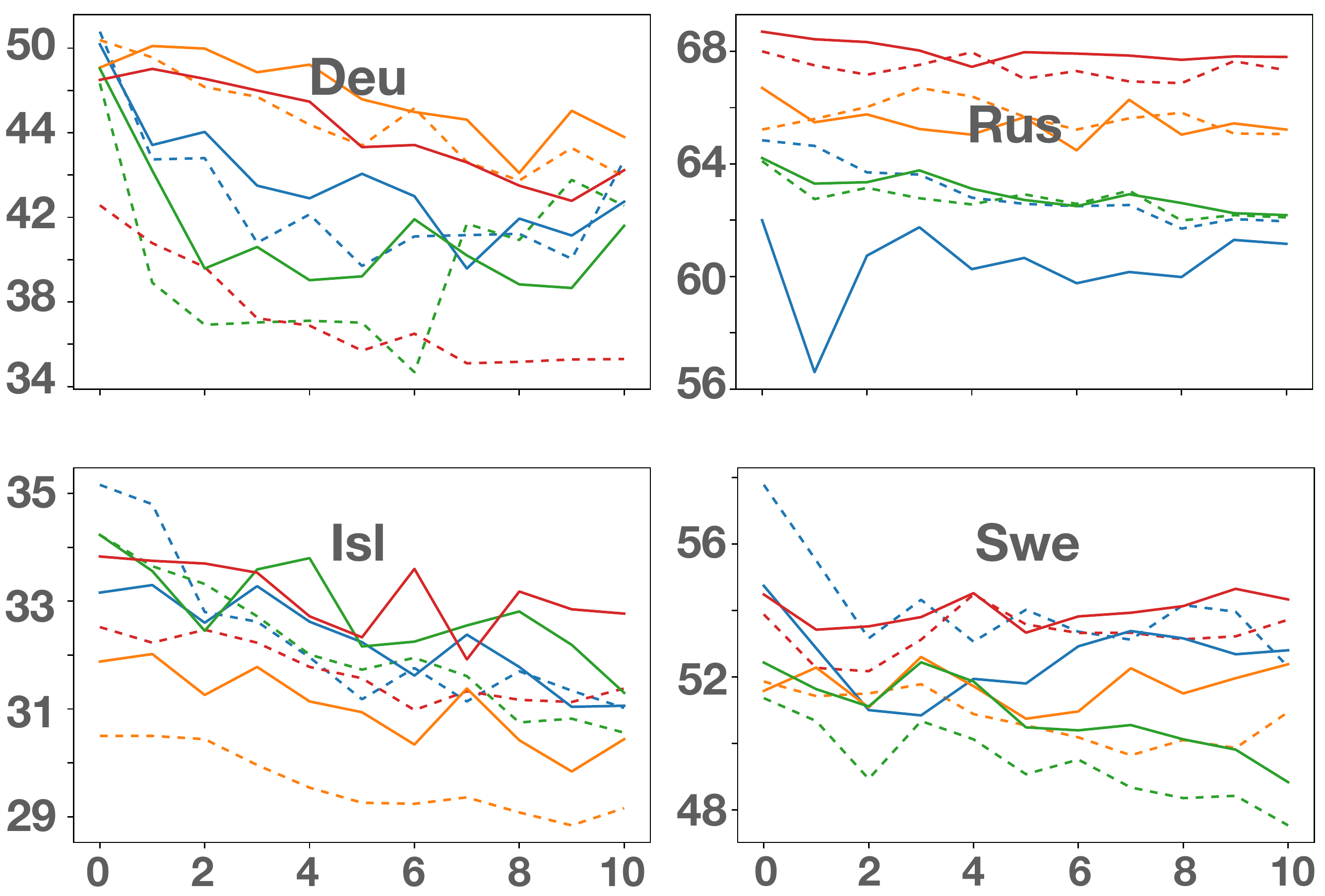
        }
        \caption{
            UniMorph
        }
        \label{fig:augmented_all_langs_unimorph}
    
    \end{subfigure}
    \caption{
        Change in accuracy as each dataset is augmented with noise. One tenth of the noisy data for a given dataset is added to the dataset at each point, until all noise is added at point 10.
    }
    \label{fig:augmented_all_langs}
\end{figure*}

\begin{figure*}[t]
     \small
     \centering
     \adjustbox{max width=0.88\textwidth}{
     \begin{subfigure}[b]{0.49\linewidth}
         \centering
         \includegraphics[scale=0.278]{
            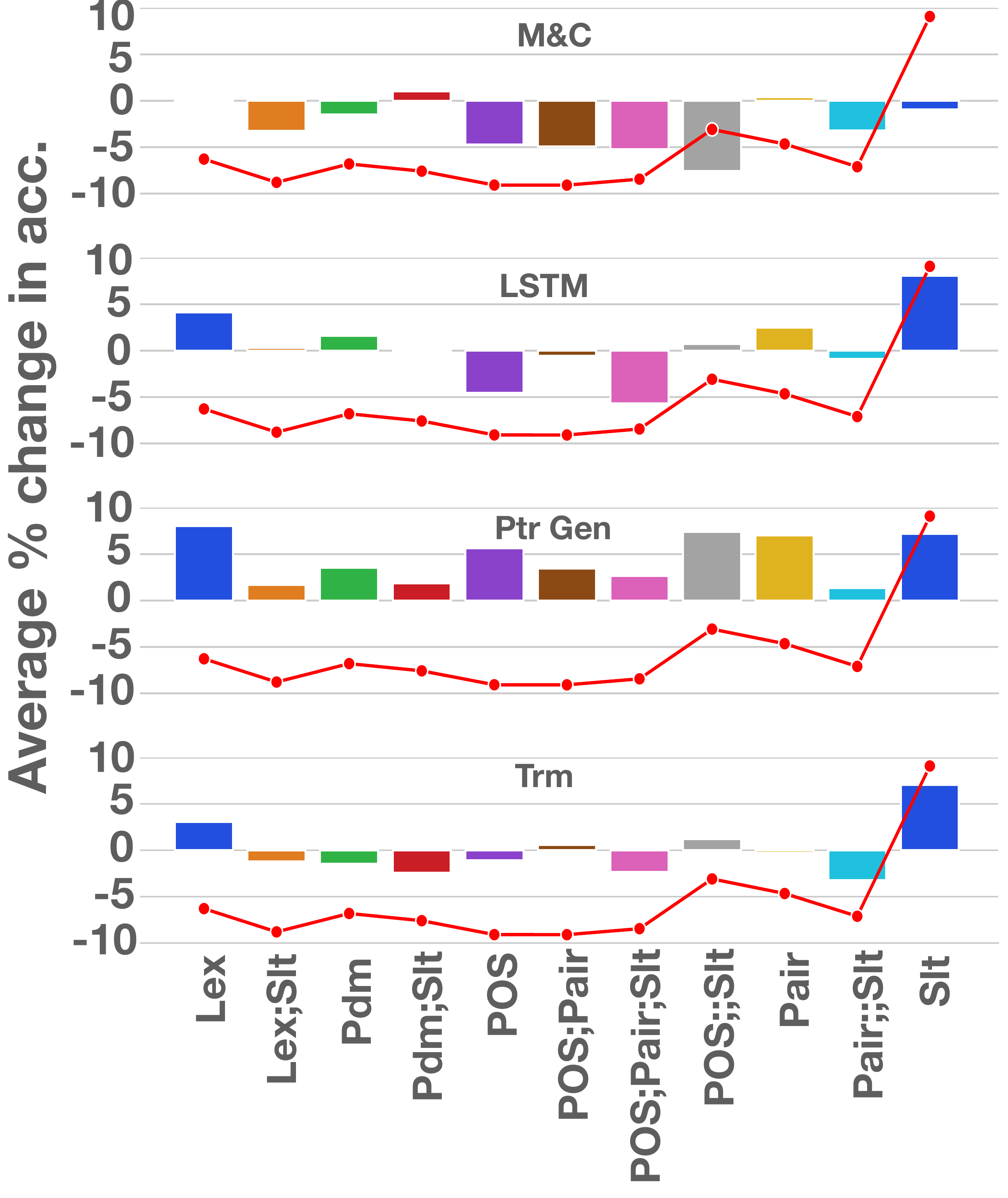
        }
        \caption{
            tUMPC
        }
        \label{
            fig:add_one_in_tumpc
        }
     \end{subfigure}
     \hfill
     \begin{subfigure}[b]{0.49\linewidth}
         \centering
         \includegraphics[scale=0.28]{
            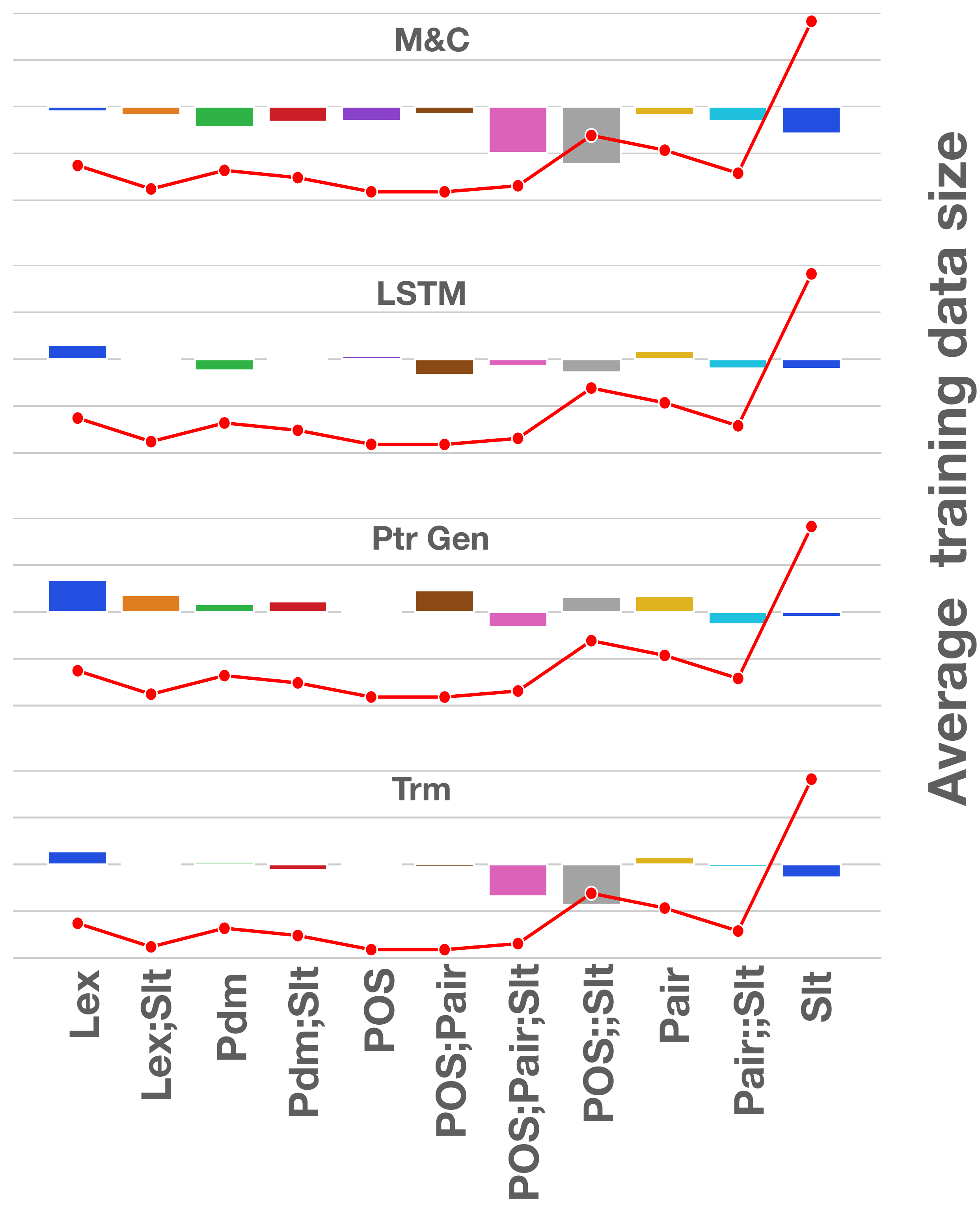
        }
        \caption{
            UniMorph
        }
        \label{
            fig:add_one_in_resampled
        }
     \end{subfigure}}
    \caption{Impact of noise by type. Each bar shows the \% increase in accuracy when samples with the corresponding annotation are added to a training set comprising only correct data. The red line represents the training data size after adding those samples.}
    \label{fig:add_one_in_exps}
\end{figure*}

\subsection{Experiment 1: Training on Noisy Data}
We first benchmark each model on all four languages when trained on the full noisy dataset.

\paragraph{tUMPC}
In Table \ref{tab:baseline_res.} we present the results when models are trained on all data that we were able to classify in our annotation pipeline, cf. Table \ref{tab:baseline_training_data}.
\makarov{} performs best on average, with \ptr{} performing second best. Notably, both of these models are designed with an inductive bias towards copying from the lemma.
\lstm{} and \trm{}, which both rely on an encoder-decoder that generates from the vocabulary at every time-step, perform similarly to one another. 
The largest training dataset, Russian, is the most accurate for every model, and Icelandic, the smallest dataset, is the least accurate.

\paragraph{UniMorph}
Table \ref{tab:resampled_res.} shows a large increase in accuracy for every language and model when compared to tUMPC. This indicates that the tUMPC data lacks diversity.
Accuracy is similar for all models on average, but \lstm{}'s accuracy is highest, and \ptr{}'s is lowest.
In a second experiment, we sample UniMorph pairs from the tUMPC word-length distribution in order to control for the fact that many UniMorph words tend to be uncharacteristically long. 
This lowers the type frequency of the dataset compared the the original UniMorph sampling --- which we interpret as reducing the diversity --- and results in a very small increase in performance compared to training on the tUMPC dataset. For results on UniMorph, we focus on the initial, more diverse dataset.

\subsection{Experiment 2: Quantity of Noise}
We investigate how introducing randomly sampled noise into a training dataset affects model performance as the amount of noise increases. This characterizes the behavior of  models as each dataset becomes noisier, and it also shows us at which quantities noise becomes most problematic.
We first train models on data comprising only samples that were annotated as correct in order to benchmark performance in the absence of noise. We then partition all of the noisy samples into ten equally sized splits.
The x-axis in Figure \ref{fig:augmented_all_langs} represents the number of partitions that have been added to a given dataset, so, at 10, we get the results in Tables \ref{tab:baseline_res.} and \ref{tab:resampled_res.}, and, at 0, we have the results when all noise is removed. 
Notice that the amount of noisy data in a given partition depends on the language, and represents one tenth of the total noise --- we focus our analysis on trends more so than performance at any particular point.
Here we analyze the dotted lines, the solid lines represent the results of Experiment 4 on CMLM. 

\paragraph{tUMPC}
Every architecture is negatively impacted by noise in German, with a huge negative impact on \makarov{}.
\ptr{} behaves sporadically, increasing performance when the last few noise partitions are introduced, and \lstm{} 
suffers the least from noise.
In Swedish, Icelandic, and Russian,
model performance tends to increase as noise is introduced. 
This is consistent in \trm{} and \lstm{} and not always true for \makarov{} and \ptr{}.

The negative impact of noise on \makarov{} in German, and the somewhat sporadic behavior of \ptr{} indicate that the classic encoder-decoder may be more robust to noise.
However, the ranking of models by accuracy does not tend to change over different amounts of noise, demonstrating that they underperform on tUMPC.
We see consistent increases in performance in Swedish for all models. One explanation for this could be that the large amount of lexicon noise in Swedish tends to contain real inflections that our models learn from.
We additionally see several other cases where \lstm{} and \trm{} increase in accuracy as more noise is added. This may simply be because they suffer from a lack of sample diversity and more data helps even though it is noisy.

\paragraph{UniMorph}
In the UniMorph dataset, \makarov{} performance steadily, and sometimes drastically, decreases as noise is introduced for every language. 
In German, \ptr{} behaves sporadically again, and both \trm{} and \lstm{} decrease in performance as noise is added, with \lstm{} decreasing very slowly.
The same is true for Swedish and Icelandic, where \trm{} and \lstm{} are less impacted by noise than the copy models.
All models besides \makarov{} seem relatively robust to noise in Russian.

This suggests that \lstm{} may be the most robust to noise in our data on average, and that the models with an inductive bias towards copying from the lemma are less robust.
In all languages but Russian, there is a distinct downward trend as noise is added for every architecture. 
Russian is the largest dataset and has more correct than noisy data. This suggests that, with sufficient correct training data, noise is less of a problem. 
Additionally, a large portion of Russian noise are slot and paradigm noise, which could feasibly have a less negative impact on learning. 
This also shows that filtering out noisy samples is most useful for \makarov{}, and the negative impact of noise is more severe for models trained on higher sample diversity.

\subsection{Experiment 3: Type of Noise} 
We investigate the impact of each noise type in our annotation schema to see if certain noise types are more important than others and if they affect architectures differently.
We produce $k$ training data sets, where each set comprises all of the correct data and all of the samples that have been annotated with one particular annotation so that we can measure that annotation's impact in isolation.
Thus, the impact of each annotation is both a function of the errors entailed by it and the frequency of samples with that annotation. 
Because noise types can co-occur on a single training sample, we consider the unique combination of noise types as a single annotation. 
Figure \ref{fig:add_one_in_exps} presents the percent change in accuracy over training on only the correct data for each dataset, averaged over all languages. 
To represent the effect of data size, we add the red line to track the average training set size when a given annotation is included.

\paragraph{tUMPC}
\makarov{}, the best performing model, is most negatively impacted by any single noise type in both the tUMPC and the UniMorph dataset, which supports the findings of the previous experiments. 
POS noise, especially in combination with slot noise has the largest negative impact. 
One explanation for this is that \makarov{} is trained to learn a policy over edits on the input, framing inflection as a true string transduction task. 
Since POS noise is likely to add inflections that are not actually meaningful, this may result in an edit policy that generates made-up words. 

Every model \textit{besides} \makarov{} gets better when the slot annotation is included in the tUMPC dataset. 
This may be due to the fact that there are a huge number of slot annotations to learn from, meaning each model is trained on a larger training set.
Additionally, \makarov{} may be less sensitive to slot errors as it has a bias towards copying from the lemma -- perhaps the best \makarov{} model trained on tUMPC ignores some parts of the MSD. 
However, \ptr{}, which also has a bias towards copying from the lemma, increases in accuracy when \textit{any} noise, especially lexicon noise, is added to the data.

\paragraph{UniMorph}
\makarov{} behaves similarly here, except the slot annotations have a negative impact. \ptr{} still learns from some noise types, but is generally less affected by any single noise type. 
\lstm{} is almost completely unaffected by any single noise type, which corroborates Experiment 2. 
\trm{} behaves similarly to \makarov{}, though the impact on accuracy is smaller. 
This indicates that with sufficiently diverse training data, \lstm{} is more robust to noise than other models, and that \trm{} learns an inductive bias that resembles \makarov{}.

\subsection{Experiment 4: CMLM Pretraining}
We experiment with a simple character masked language modeling (CMLM) pretraining objective as a method for improving noise-robustness of each model. 
Auto-encoding without masking has been shown to impart a helpful inductive bias towards copying from the lemma in low-resource morphological inflection \cite{kann-schutze-2017-unlabeled}.
Masking can be thought of as additionally adding a denoising objective to auto-encoding, which may contribute to more robust learning from noisy training samples. 
We experiment with CMLM for every model and analyze its effect on noisy training.

\begin{figure}[t]
     \centering
     \includegraphics[scale=0.20]{
        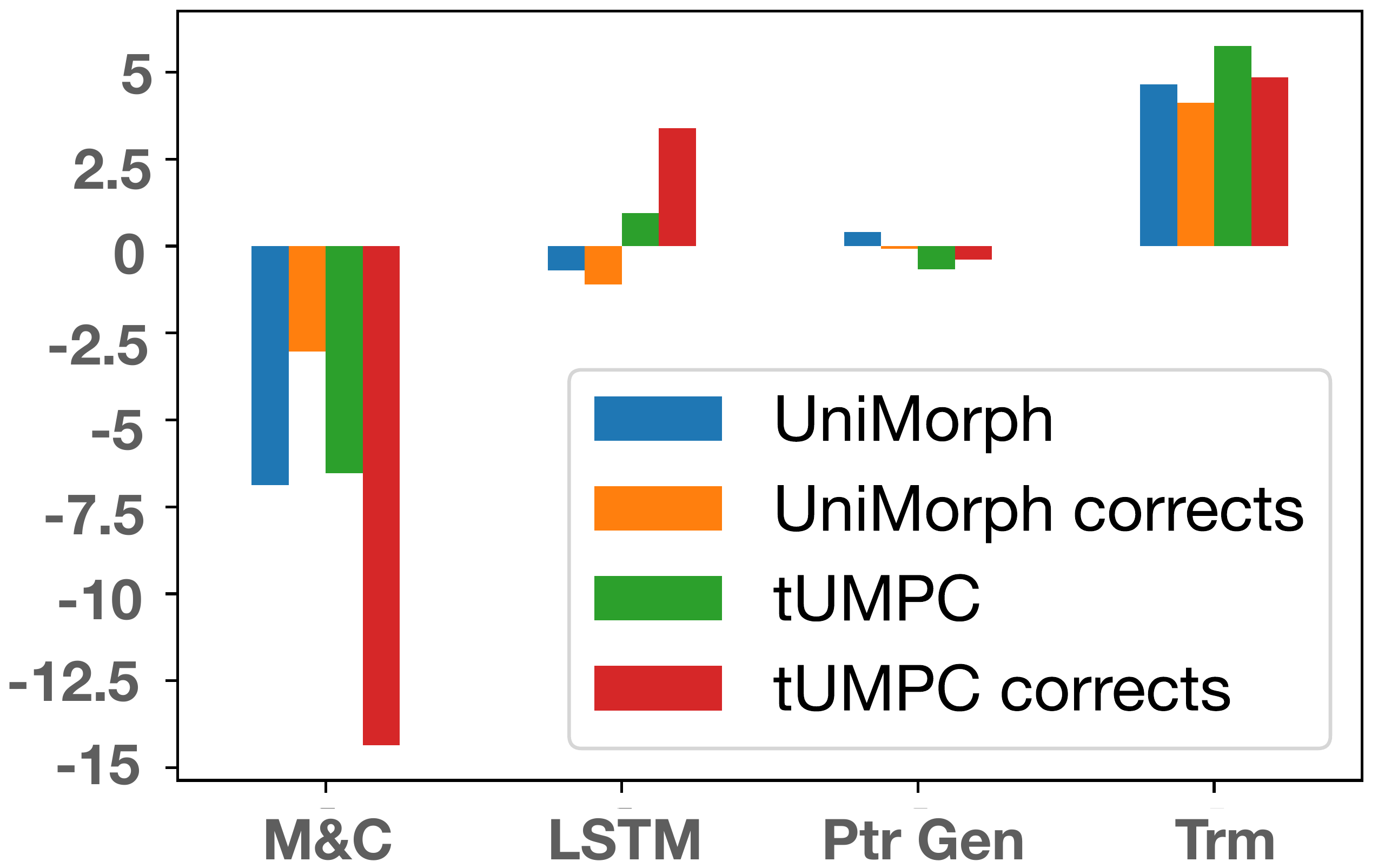
    }
    \caption{
        \% change in accuracy from pretraining.
    }
    \label{fig:pretraining_rel_diff}
\end{figure}
\paragraph{Implementation}
We follow the BERT masked language modeling objective \cite{devlin-etal-2019-bert} with only minor adjustments. 
First, we mask characters rather than subwords. Second, our sequences will be shorter on average than BERT, so we increase the mask probability from 0.15 to 0.2. 
In practice, we follow RoBERTa \cite{liu2019roberta} and generate a new mask for each sample dynamically at each epoch.
Every unique type in a given training set, without their MSDs, comprise the pretraining dataset. 
We do not include additional types in order to test whether CMLM itself is effective, rather than the addition of data.
Each model is otherwise trained in the exact same way as finetuning with the same hyperparameters other than number of epochs. 
The only exception is that we train \ptr{} with a warm-up scheduler in order to avoid over-fitting to copying the lemma. Exact hyperparameters are in the appendix in Table \ref{tab:appendix_pretrain_hyperparams}.

\paragraph{Results}
Figure \ref{fig:pretraining_rel_diff} shows the average percent change in accuracy for each model in both datasets when adding the pretraining objective. 
We find that it has little effect for \ptr{} on average, and actually has a negative impact on \makarov{}. 
We attribute this to the fact that these models have an inductive copy bias, which may be a large part of what CMLM adds to the models. 
\makarov{} might additionally be negatively impacted if the imitation learning objective overfits to the edits during pretraining. 
More experimentation with pretraining objectives are needed to understand this negative result, however. 
Our focus is on the impact of pretraining on noise robustness.

\lstm{} is barely affected by pretraining on average. 
\trm{}, however, increases in accuracy in both datasets when pretrained. It additionally increases in accuracy when training on correct data only, suggesting that CMLM may be generally beneficial to \trm{}, not just in noisy settings. 
We note, however, that the increase in accuracy is greater when noise is included in both datasets.

\paragraph{CMLM by Noise Quantity}
We also reproduce Experiment 2 in order to look more closely at the impact of pretraining. 
We focus on the solid lines in Figure \ref{fig:augmented_all_langs}, and compare them to the dotted lines that represent the models that were not pretrained.

\paragraph{tUMPC}
For \makarov{}, we see the negative impact of pretraining is especially strong in German and Swedish, and pretraining has little effect in other languages.
Pretraining benefits all other architectures in Swedish and Icelandic, especially as noise is added. In German and Russian, pretraining is not beneficial to \ptr{}, but still helps \trm{} and, at small amounts of noise, \lstm{}. 
Pretraining is also beneficial when there is no noise in the dataset, but to a lesser extent and not in every language.

\paragraph{UniMorph}
\makarov{} shows performance increases in German under small amounts of noise only. 
But in every other language pretraining still impacts \makarov{} negatively.
Similar to the tUMPC data, pretraining benefits all other models in Icelandic and Swedish.
Even without noise, \lstm{} and \trm{} increase in performance in Icelandic and to some extent in Swedish and Russian. 
In German, \trm{} increases massively in accuracy from pretraining, including when there is no noise. 
All other architectures also benefit from pretraining in German, but largely only at small amounts of noise. 
Still, pretrained \lstm{} is the best overall model for German in the UniMorph dataset. 
In Russian, \trm{} is the only pretrained architecture that performs better as noise is introduced. Pretraining benefits \lstm{} in Russian noise when there is no noise, however.

\trm{} is the only architecture for which CMLM pretraining helps in every language and dataset. 
Overall, pretraining is also clearly beneficial to \lstm{}, and sometimes to \ptr{}. 
Pretraining is particularly effective for these three architectures in Swedish and Icelandic, especially as noise is introduced to the datasets.
Combined with the fact that these two languages have more noisy than correct data, this implies that pretraining \textit{is} effective for noise-robust learning.
Still, in many cases pretraining leads to a gain in accuracy when there is no noise in the dataset, implying that this learning strategy is generally beneficial, especially for \trm{}.

\section{Discussion}
We find that low sample diversity has a strong impact on performance of all models. 
The tUMPC training setup favors architectures with a copy bias, and demonstrates that models can learn from noisy training samples when the dataset is not diverse.
We find that the low \lstm{} and \trm{} performance is largely explained by low sample diversity. 
On the other hand, they seem to be more robust to noisy data, particularly \lstm{}, which has stable performance as noise is added to the UniMorph dataset.
Pretraining with CMLM leads to further gains in performance for \lstm{}, but the largest gains from CMLM are for \trm{}. 
On average, \trm{} pretrained with CMLM is the best performing model under noise, when we have sufficient sample diversity.
When we look at specific noise in the data, we find that slot alignment issues in tUMPC tend to have low impact on every model. 
This could be for several reasons: slot noise should be irrelevant under high amounts of syncretism, which is abundant in German, 
Icelandic, and Swedish. Additionally, as models become more corrupted by noise, they may rely on a bias towards copying from the source form -- which is unaffected by the slot.

We find that certain noise that results from errors in paradigm induction are particularly impactful for \makarov{}. 
This is especially true for POS noise, which may motivate better POS induction in unsupervised morphology systems. Under greater sample diversity, \trm{} is similarly impacted by paradigm induction errors.
This suggests that \trm{} begins to learn an inductive bias similar to \makarov{} on noisy data.
The addition of almost any single noise annotation leads to an increase in accuracy in \ptr{}, and removing noise from the training data often impacts \ptr{} negatively. 
However, Experiment 2 suggests that \ptr{} is not as robust to increasing amounts of noise as other models.
A lot of reduction in accuracy may come from combinations of different noise types, which is not captured by Experiment 3. 
Future work could investigate noise distributions by type with particular focus on \ptr{} behavior.
We additionally find that \makarov{} is negatively impacted by CMLM pretraining. We believe this may be due to overfitting its copy bias and learning spurious transductions from the masking objective. However, future work should consider alternate pretraining strategies for \makarov{}.

Overall this implies that, although copy models are preferred for training on low sample diversity, classic encoder-decoders are a good choice for noisy datasets with more diversity. 
Our results indicate that POS and paradigm induction components are more important for training data quality than slot alignment in unsupervised systems and that bootstrapping inflection pairs should prioritize lemma diversity, even if it may induce noise.

\section{Conclusion}
We have investigated the impact of noise on state-of-the-art neural morphological inflection models. We find that the noise that arises in an unsupervised system for bootstrapping inflection pairs is frequently related to slot alignment errors, but that those also have less impact on the models. We have also compared two inflection architectures with a copy bias to two typical encoder-decoder models. We find that, though copy bias is helpful under low sample diversity, the encoder-decoders are more robust to noise.
Finally, we find that a simple masked pretraining objective makes encoder-decoders, and especially Transformers, more accurate under noise.

\section{Limitations}
The largest limitation of this study is that our annotation pipeline is automated. This makes it possible that there are errors in the noise annotations that we base our analysis on. Additionally, since we capture a naturally occurring noise distribution, our findings are coupled to the datasets we study here. Our findings may not generalize to distributions of noise in other datasets.

\section*{Acknowledgments} We thank the anonymous reviewers for their helpful feedback.

\bibliographystyle{acl_natbib}
\bibliography{anthology,custom}

\newpage
\appendix

\section{Appendix}
\label{sec:appendix}

\begin{table*}[t!] 
\centering
\small
\begin{tabular}{l|cccccc|ccc} 
\toprule 
Language & C & LEX & PDGM & POS & POS Pair & SLOT & Lemma Overlap & MSD Overlap & Tag Overlap \\
\toprule
deu & 61.37 & 1.52 & 0.68 & 8.07 & 8.60 & 19.75 & 3.84 & 24.32 & 83.33 \\
isl & 37.95 & 3.91 & 2.11 & 15.59 & 8.17 & 32.27 & 3.24 & 60.61 & 95.65 \\
rus & 50.77 & 0.98 & 6.01 & 4.21 & 4.25 & 33.79 & 3.31 & 68.18 & 100.0 \\
swe & 27.90 & 11.36 & 5.48 & 8.39 & 15.66 & 31.20 & 1.94 & 69.33 & 93.1 \\
\bottomrule
\end{tabular}
\caption{Statistics for each language's training data. The \% of samples with a given annotation (left). The \% overlap with the evaluation set for lemmas, MSDs, or individual tags (right).} 
\label{tab:training_data_stats.} 
\end{table*}

\begin{table}[t!] 
\small
\centering
\begin{tabular}{l|lllll} 
 \toprule
 Isl & verb & adjective & noun & & \\
 Deu & verb & adjective & noun & prn & \\
 Rus & verb & adjective & noun & prn & numeral \\
 Swe & verb & adjective & noun & & \\
\bottomrule
\end{tabular}
\caption{Parts of speech that inflect in our annotation schema.} 
\label{tab:appendix_valid_pos} 
\end{table}

\begin{table}[ht]
\small
\centering
\setlength{\tabcolsep}{4pt}
\begin{tabular}{l|llll}
\toprule 
 & \makarov{} & \ptr{} & \lstm{} & \trm{} \\
 \midrule
 Epochs & 60 & 60 & 60 & 800 \\
 Batch size & 1 & 32 & 20 & 400 \\
 Optimizer & AD & adam & AD & adam \\
 LR & 1.0 & 0.001 & 1.0 & 0.001 \\
 Scheduler & - & - & - & inv. sqr root \\
 Warmup & - & - & - & 4000 \\
 Total Params & 260k & 1.1M & 370k & 7.4M \\
\bottomrule
\end{tabular}
\caption{Hyperparameters for each architecture. All other hyperparameters follow their respective publications exactly. \textit{AD}=ADADELTA.} 
\label{tab:appendix_hyperparams.} 
\end{table}

\begin{table}[ht] 
\centering
\small
\setlength{\tabcolsep}{2pt}
\begin{tabular}{l|llll} 
\toprule 
 & \makarov{} & \ptr{} & \lstm{} & \trm{} \\
 \midrule
 Pretrain epochs & 40 & 40 & 40 & 200 \\
 Pretrain scheduler & - & inv. sqr root & - & inv. sqr root \\
 Pretrain warmup & - & 100 & - & 1000 \\
\bottomrule
\end{tabular}
\caption{Hyperparameters for each architecture during pretraining.} 
\label{tab:appendix_pretrain_hyperparams} 
\end{table}

\subsection{Slot Mapping Details}
\label{sec:appendix_slot_map}
We begin with each type processed by tUMPC, which has a slot: an arbitrary identifier for its POS and inflection category, and a (not disambiguated) morphological analysis from Apertium. For example, given the German verb \textit{tragt}, we have a tUMPC slot $2$. We additionally have the analysis from Apertium with two possibilities: an imperative plural verb (\textrm{<vblex><imp><pl>}) or a second person present indicative plural verb (\textrm{<vblex><pri><p2><pl>}). 

Each possibility in the Apertium analysis is then mapped to a UniMorph MSD via a mapping we create that translates each tag one at a time. For example, the tag <vblex> becomes V, in order to match the UniMorph schema. After some language specific post-processing heuristics, we get a set of UniMorph MSDs from every Apertium analysis.

We can use these mapped analyses to align tUMPC slots with UniMorph MSDs. Consider our example above, \textit{tragt}, where we would end up with two possible MSDs: V;IMP;2;PL and V;IND;PRS;2;PL. This forms a mapping from the slot $2$, to both of these MSDS. Due to tUMPC errors, we may also erroneously get a mapping from slot $2$ to N;ACC;PL via some other word. This gives us a mapping from three differing UniMorph MSDs to one tUMPC slot. Over all such mappings, this forms a bipartite graph between tUMPC slots and UniMorph MSDs, where the same UniMorph MSD may correspond to multiple tUMPC slots. However, one tUMPC slot represents exactly one MSD. We thus follow \citet{jin2020unsupervised} and attain a one-to-one mapping from tUMPC slots to MSDs by finding the matching that maximizes the overlap of word types with an aligned slot and MSD. Like them, we use the algorithm from \citet{karp1980algorithm} to optimize this matching. Finally, the slot for every training sample can be mapped to it's MSD according to this matching. 

\subsection{Noise Detection}
\label{sec:appendix_detecting_noise}
Here we describe how our annotation pipeline detects each type of noise. We rely on Apertium for the entire pipeline. Though most noise is found with original Apertium analysis, the analyses that have been mapped to UniMorph MSDs are used for detecting slot noise.
.
\paragraph{Lexicon Noise} 
We use three resources to form the lexicon of each language: Apertium, Wikipedia, and a python spellchecker\footnote{\href{https://github.com/dvwright/phunspell}{https://github.com/dvwright/phunspell}} based on hunspell\footnote{\href{http://hunspell.github.io/}{http://hunspell.github.io/}}. Any word not in any of these three is lexicon noise.

\paragraph{POS Noise}
Here we produce lists of POS that inflect for each language. Any word whose Apertium analysis does not contain any valid POS according to this list is considered POS noise. The valid POS for each language are listed in Table \ref{tab:appendix_valid_pos}.

\paragraph{POS Pair}
Here we consider all POS from Apertium analysis of both words in a pair. If they have no POS in common, then it is POS pair noise.

\paragraph{Paradigm}
For a given inflection pair, if both words have no overlapping lemmas of the same POS (but \textit{do} have a shared POS) according to Apertium, we consider them to be from separate paradigms, and thus paradigm noise.

\paragraph{Slot}
Slot noise occurs when the slot assigned to a word by tUMPC is not in the set of slots from Apertium analysis. We rely on the version of tUMPC and Apertium slots that have been mapped to UniMorph for this part of the annotation. Slot noise considers only the predicted and gold MSD for a slot, so it can can co-occur on a single sample with any other noise in the taxonomy.

\end{document}